# Turn Down that Noise: Synaptic Encoding of Afferent SNR in a Single Spiking Neuron

Saeed Afshar, *Member IEEE*, Libin George, *Member IEEE*, Jonathan Tapson, *Member IEEE*, André van Schaik, *Fellow IEEE,* Philip de Chazal, *Member IEEE* and Tara Julia Hamilton, *Member IEEE*

Biomedical Engineering and Neuroscience Program
The MARCS Institute
University of Western Sydney
Kingswood, NSW Australia
s.v.afshar@uws.edu.au

*Abstract*—We have added a simplified neuromorphic model of Spike Time Dependent Plasticity (STDP) to the Synapto-dendritic Kernel Adapting Neuron (SKAN). The resulting neuron model is the first to show synaptic encoding of afferent signal-to-noise ratio in addition to the unsupervised learning of spatio-temporal spike patterns. The neuron model is particularly suitable for implementation in digital neuromorphic hardware as it does not use any complex mathematical operations and uses a novel approach to achieve synaptic homeostasis. The neuron's noise compensation properties are characterized and tested on noise corrupted zeros digits of the MNIST handwritten dataset. Results show the simultaneously learning common patterns in its input data while dynamically weighing individual afferent channels based on their signal to noise ratio. Despite its simplicity the interesting behaviors of the neuron model and the resulting computational power may offer insights into biological systems.

I. Introduction

Synapses are by far the most numerous computational elements in our spiking brains and in spiking neuromorphic systems. Due to their large numbers, the return on investment on synapses, i.e., how much functional computation they perform versus how much hardware resources they take up becomes a defining feature of any neural system whether evolved or designed by an engineer [1][2][3].

Leaving aside the stream of neuromorphic engineering which aims to create biologically accurate models of neurons in silicon, the extraction of the most functionality from the fewest, simplest synapses is often a central focus for the neuromorphic engineer [4][5].

In the context of neuromorphic systems the synapse serves three essential functions. The first is simply to form a connection from one neuron to the next. The second is to spread the energy of the presynaptic input spike over time via the synaptic kernel and the third is to weigh this kernel such that when it is added to other similarly weighted synaptic kernels, the resulting summation, called the somatic membrane potential, is a 'good' signal that encodes functionally useful information.

While the realization and use of these distributed kernels based processing units has evidently been mastered by evolution, despite significant recent progress, our best engineered systems still find the large-scale realization of these three functions challenging.

The first and simplest function of the synapse, that of acting as the connection between neurons represents the greatest hardware challenge. Limitations in network bandwidth or connectivity are often the most serious obstacles restricting full utilization of neuromorphic hardware resources. Innovative approaches such as time multiplexing [3] and AER [6][7] can create virtual all-to-all connected networks, however, these advantages come at the expense of reduced operating speed.

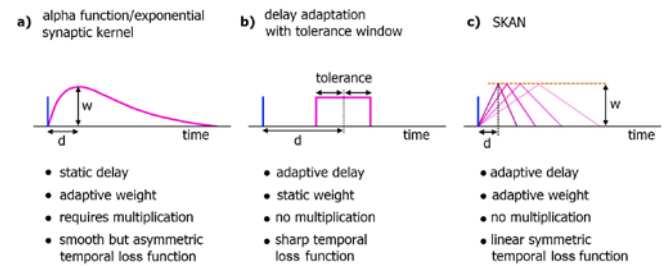

Figure 1. Comparison of neuromorphic implementations of synapto-dendritic kernels. The characteristics of realized EPSP kernels are computationally important just prior to being summed at the soma. These kernels represent the loss function used to translate the temporal error in spatiotemporal spike patterns at the synapse to the somatic membrane potential. Due to the large number of synapses neural network systems require, the complexity, functionality, and hardware cost of these kernels is a critical feature of neuromorphic spiking networks.

The second function of the synapse, that of spreading an input spike's energy over time can be realized via a range of synaptic kernels with varying levels of complexity, hardware cost and computational utility. Two extremes include the biologically plausible exponentially decaying synaptic alpha function shown in Figure 1 (a) and the simple delay learning system with a binary kernel and a temporal tolerance window shown in Figure 1 (b). The canonical synaptic alpha function is multiplied by a real valued synaptic weight to model synaptic weight adaptation. The costs of implementing this synaptic kernel in digital hardware however are substantial. The kernel requires at least two real valued multiplication operations at each time step for each synapse (one for the kernel and one for the weight). This makes the kernel unsuitable for implementation in distributed digital hardware. In addition to being costly, the alpha function does not model the computationally useful peak delay adaptation effects observed in biology [8].

At the other end of the spectrum, rather than implement complex synaptic weight adaptation, other neuromorphic SNN implementations have focused exclusively on adjustment of explicit propagation delays along the neural signal path to encode memory [9][10][11]. Here the energy of the spike is spread via a binary valued tolerance window as shown Figure 1 (b). This discarding of synaptic weights significantly simplifies implementation and allows more synapses to be realized. The down side is that explicit window-based delay learning schemes can produce "sharp" systems with lower tolerance for the dynamically changing temporal variance they inevitably encounter in applications where neuronal systems are expected to excel: noisy, dynamic and unpredictable environments [12]. In addition while use of these simplified kernels allow more synapses to be realized, limited network bandwidth can sometimes mean that these larger numbers cannot actually be fully utilized.

Against this background, the proposed SKAN model uses fully adaptable yet simple kernels with more functionality per synapse such that fewer synapses (and thus fewer connections) are required enabling better use of available hardware and bandwidth resources while still providing tolerance to noise [13]. SKAN's live unsupervised hand gesture learning and recognition has been in [14] using a neuromorphic visual scene to spatio-temporal spike pattern transformation [15].

## II. SYNAPTO-DENDRITIC KERNEL ADAPTATING NEURON

### A. Kernel and threhsold adaptation mechanisms

The dynamics of SKAN have been previously described in detail for the special case of static weights [16]. Figure 2 illustrates the functional diagram of the neuron.

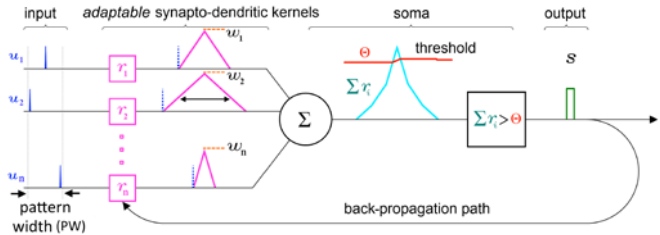

Figure 2. Schematic of the elements and information paths in SKAN. The input spikes (blue) trigger adaptable synapto-dendritic kernels (magenta) which rise up to the synaptic weight (orange) and are summed to form the neuron's somatic membrane potential (cyan). This is then compared to an adaptive somatic threshold (red) which, if exceeded, results in an output pulse (green). The output pulse also feeds back to adapt the kernels. Note that the back propagating signal does not travel beyond the synapto-dendritic structures of the neuron to previous neural layers.

At each synapse the presynaptic input spike $u_i(t)$ triggers the EPSP kernel, which is the central element of SKAN and is modeled as an adaptable kernel in the form of a simple ramp-up-ramp-down sequence $r_i(t)$. After being triggered $r_i(t)$ rises with slope $\Delta r_i(t)$ until it reaches the synaptic weight variable $w_i(t)$, after which the kernel ramps down with the same slope and returns to zero. These EPSP kernels spread the energy of the input spike signals over time allowing them to be summed at the soma to generate the somatic membrane potential $\Sigma r_i(t)$, which is compared to the threshold $\Theta(t)$ to generate an postsynaptic output pulse $s(t)$. The back propagation of this output pulse in turn adapts the slope of the kernels $\Delta r_i(t)$ such that, if a kernel is rising at the time of the output pulse, it is deemed to be too late and so its slope is increased, making the kernel sharper. Alternatively if the kernel is falling at the time of the output pulse, it is deemed to be too early and its slope is decreased such that the kernel becomes wider. This kernel slope adaptation mechanism aligns the peaks of many synaptic kernels in response to repeated presentations of the same spatio-temporal pattern. As the kernel peaks become ever more aligned their summation at the somatic membrane potential forms an ever higher and narrower peak as shown in Figure 3.

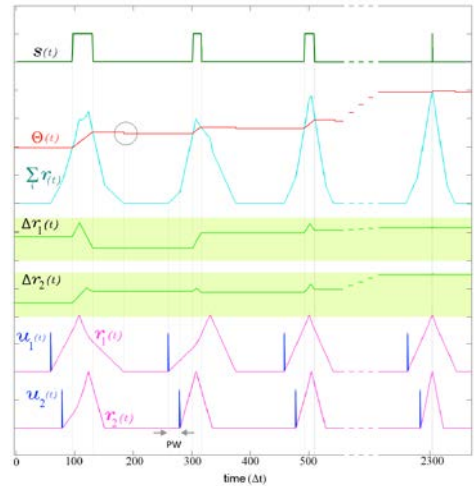

Figure 3. Kernel adaptation in SKAN with static synaptic weights. The kernels and the threshold of SKAN adapt in response to repeated spatio-temporal pattern presentations. The kernels have captured the ISI information by the third presentation of the pattern. With each subsequent presentation the threshold $\Theta(t)$ increases making the neuron more selective as the kernel step sizes $\Delta r_i(t)$ increase, making the kernels narrower. As a result, each pattern presentation increases the neuron's confidence about the underlying process producing the ISI's, narrowing the neuron's receptive field around the target ISI, and producing a smaller output pulse $s(t)$. By the 11[th] presentation ($t=2300\ \Delta t$), the $\Theta_{rise}$ during the output spike and $\Theta_{fall}$ balance each other such that the $\Theta_{before} \approx \Theta_{after}$. The soma output spike $s(t)$ is now a finely tuned unit delta pulse which indicates high certainty. When the membrane potential returns to zero, the neuron's threshold falls as indicated by the grey circle.

In addition to the kernel adaptation, which captures the temporal information of the input pattern, the neuron's threshold also adapts such that during an output pulse the threshold rises and every time the membrane potential returns to zero the threshold falls, as indicated by the grey circle in Figure 3. As the neuron spikes more in response to a particular pattern, its threshold rises, making the neuron ever more selective for the pattern that triggered it and narrowing its spatio-temporal receptive field. Conversely, unrecognized input patterns, which do not cause an output spike, reduce the threshold, making the neuron more receptive to new patterns. Through this feedback mechanism the neuron automatically maintains a balance between selectivity and generalization in response to the statistics of its environment.

In the following sections the synaptic weight adaptation of SKAN is described and the resulting behaviors of the neuron are demonstrated.

## III. Weight Adaptation via Simplified STDP

### A. Simplified Weight Update Rule

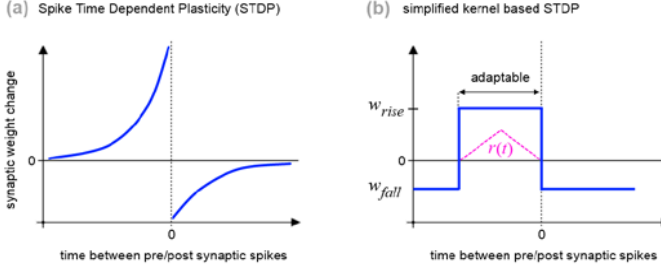

Figure 4. Comparison of the classical STDP synaptic update curve (a) with the simplified kernel based STDP used in this work (b) where in addition to learning the spatio-temporal pattern, the adaptable EPSP kernel also doubles as a flag that enables the increase of synaptic weights in the event of a postsynaptic spike.

In biology the rules governing synaptic weight adaptation vary enormously both in degree and in type across species, brain regions, synapse types, cell types, within individual cells, over short time scales, and as a function of organism development [17]. However, the STDP rule shown in Figure 4 (a) is by far the most studied synaptic plasticity rule in neuroscience today due to its reproducibility and neurocomputational utility. As a consequence, the faithful modeling of this rule in hardware is now a major focus in neuromorphic engineering [18]. As with the smooth synaptic alpha function, such accurate modelling of neurobiological processes can incur additional hardware costs while providing little computational improvement compared to even highly simplified models [19]. Therefore, as was the case with SKANs simplified kernels, in this report, the classic STDP rule is replaced with the simiplified weight update rule shown in Figure 4 (b). The rule is designed so that it reuses the same signals and flags that are already present in the static weight SKAN system, such that the presynaptic input spike $u_i(t)$, which triggers the EPSP kernel $r_i(t)$, also triggers a binary weight adjustment flag $d_i(t)$. If a back-propogating postsynaptic output spike, $s(t)$, arrives while this flag is high, then the synaptic weight $w_i(t)$ is increased by $w_{rise}$ and the flag is reset to zero. Alternatively if the membrane potential $\Sigma r_i(t)$, returns to zero before an output spike arrives, then the synaptic weight is decreased by $w_{fall}$ and the flag is returns to zero. These two rules are described in Equation 1 and Equation 2.

$$w_i(t) = \begin{cases} w_i(t-1) + w_{rise} & \text{IF} \quad d_i(t-1) \wedge \downarrow s(t) \\ w_i(t-1) - w_{fall} & \text{IF} \quad d_i(t-1) \wedge \downarrow \Sigma r_i(t) \end{cases} \quad (1)$$

$$d_i(t) = \begin{cases} 1 & \text{IF} \quad u_i(t) \\ 0 & \text{IF} \quad \downarrow \Sigma r_i(t) \vee \downarrow s(t) \end{cases} \quad (2)$$

Where $\downarrow s(t)$ is the falling edge of the postsynaptic output spike and $\downarrow \Sigma r_i(t)$ is the return of the membrane potential to zero ($\Sigma r_i(t-1) > 0 \wedge \Sigma r_i(t) = 0$).

As a result of Equation 1 and Equation 2, every time the membrane potential rises due to input spikes, the synaptic weights of the activated channels either rise in response to the resultant output spike or they fall when the membrane potential returns to zero.

This simplification of the STDP model significantly reduces hardware costs. By using the EPSP kernel value as a binary flag of adaptable duration, the need for realization of an exponentially decaying signal is eliminated and the use of the constant update terms $w_{rise}$ and $w_{fall}$ replaces the addition of two arbitrary real numbers, $w_i(t)$ and $\Delta w_i(t)$, which would otherwise be required at each synapse and which is significantly more costly in terms of hardware resources in comparison to constant terms which can be hardwired.

It is important to note that, unlike in classical STDP, in the proposed model, if an output spike were to be triggered somehow after the membrane had returned to zero (say by an external stimulator) there would be no change to the synaptic weight, but in the normal operation of the neuron this reduction in the model does not affect the system's performance. Similarly, in the SKAN model multiple input spikes that arrive within a short time or in bursts are 'covered' by the EPSP kernel of the leading spike and are invisible to the system. This reduction is arguably desirable as it correlates well to real world event driven stimuli where relative stimulus onset times across afferents carry salient information.

### B. Synaptic Weight Normalization Without Division

An additional layer of complexities arises through the need for synaptic homeostasis which is required to keep the synaptic weights within some dynamic range. In the context of biology a range of synaptic homeostasis models have been presented.

Unfortunately the hardware implementation costs of the previously proposed models of synaptic homeostasis are prohibitively high.

In this work we propose a novel cheap digital approach to this problem that eliminates the need for multiplication, division or other complex operation.

Instead of normalizing the synaptic weights such that the max of the weights, max($w_i(t)$), or the sum of the weights, $\Sigma w_i(t)$, is clamped to a specific value, this summation or max signal is allowed to roam within the top half of a digital range updated by the weight update rule of Equation 1 and Equation 2. When the update rule pushes the max($w_i(t)$) or $\Sigma w_i(t)$ signal beyond this digital range, all the neuron's signals, i.e., its weights $w_i(t)$, its EPSP kernels $r_i(t)$, and its threshold $\Theta(t)$, are right shifted, which halves their values. Conversely if the signal falls below half its range all the signals of the neuron are left shifted. These two conditions simplify to Equation 3 and Equation 4.

$$\bigcup_i overflow(w_i(t)) \rightarrow right\ shift\ neuron\ values$$

(3)

$$\bigcap_i MSB(w_i(t)) = 0 \rightarrow left\ shift\ neuron\ values$$

(4)

The fact that the other neuronal parameters, $r_i(t)$ and $\Theta(t)$ also shift means that the neuron is essentially not affected by the shift operations.

An important edge case occurs in the 'right shift neuron values' operation, which requires a design decision in terms of any weak synapses which go to zero. The simplest option is to not allow any weight to go to zero. This can be implemented either by checking all bits of every synapse or simply by assuming the LSB of all synapses is set to high without any zero checking. The second option introduces a small amount of additional noise but as shown in Figure 5 the added noise is negligible.

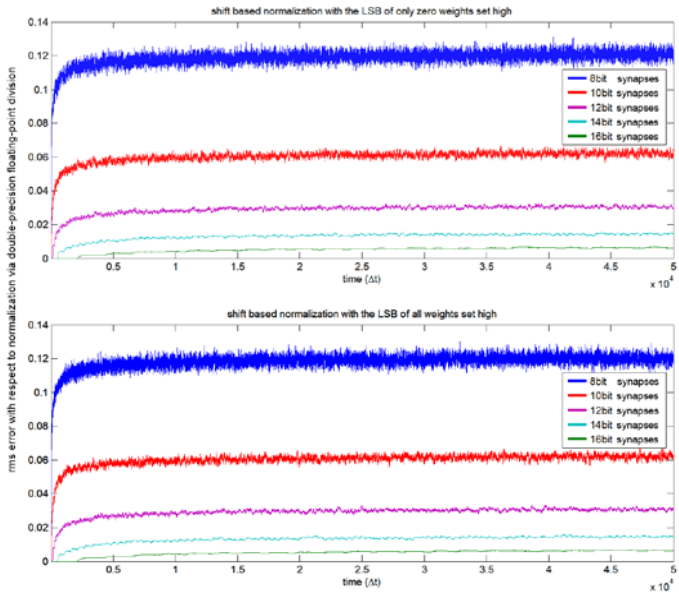

Figure 5. Comparison of shift based normalization with respect to normalization via double-precision floating-point division. Random synaptic weight updates were performed on a 16 synapse neuron of varying synaptic bit-length. The rms error of the relative weights of the shift based synapses was calculated against normalization via double-precision division. Increasing bit-length in the shift based synapses resulted in lower error but even at the lower bit-lengths the *relative order* of the synaptic weights followed the floating-point implementation. The noise introduced by setting all LSBs to high (bottom) was negligible with respect to only setting the LSB of zeroed synapses to high (top).

Another option in dealing with weights that go to zero is to disable them or potentially re-allocate their resources to other neurons. A potential application of the second option is discussed in section IV.

Thus the overall effect of the shifting operations described is to continuously 'generate more dynamic range' such that all weights become normalized while the max signal remains within the range described by Equation 5

$$2^{b-1} < \max(w_i(t)) < 2^b - 1$$

(5)

where $b$ is the number of bits used to represent the synaptic weights.

Another significant design question is *when* to normalize. Ideally the weight normalization should occur without affecting the regular learning processes outlined in the update equation. The most straight forward option is to periodically push the system into an standby state where if all is quiet (with no incoming inputs), the system slips into a weight normalization state, where the normal learning process is shut down and synaptic weight information is transmitted to the soma.

In a hardware model such an operation is simple since the epsp channel connecting the synapse to the soma has the same bit width as the synaptic weight, the weights can all be simultaneously transmitted via the same channels to the soma where the max($w_i$) or $\Sigma w_i(t)$ signals can be calculated and used for normalization. Remarkably the simple steps described have direct analogs in sleep, where synapses that have been potentiated due to learning during wakefulness, are all simultaneously activated during slow wave sleep, such that they effectivtly report their total strengths (the $\Sigma w_i(t)$ signal) to their soma. This total activation signal can then be used to gradually reduce the excitation of all synapses [20].

IV. RESULTS

*A. STDP and SKAN combine to produce Synaptic Encoding of Afferent Signal to Noise Ratio*

As demonstrated in [16] given static synaptic weights $w_i=w_0$, the simple kernel adaptation of SKAN can perform unsupervised learning of common spatio-temporal patterns in noisy environments. When this static weight model of SKAN is combined with the synaptic weight update and the normalization operation of the previous section, the neuron not only finds and learns the most common spatio-temporal pattern, but additionally adjusts its synaptic weights independently to compensate for the signal-to-noise ratio of individual input channels and thus improves recognition performance.

To demonstrate this effect, consider the case where the neuron is presented with repeated spatio-temporal spike patterns that are received via noise corrupted channels. After several pattern presentations the neuron's kernels 'see through' this noise and adapt their slopes $\Delta r_i(t)$ so that they align with the pattern. This is because the noise is uncorrelated with the pattern and it is just as likely to increase the slopes as it is to decrease them and the noise quickly gets averaged out, leaving only the target pattern for the kernels to train on. Figure 6 shows what happens next for a case where one of the three channels is corrupted by SNR=1:1, that is, where the probability of the presence of a Poisson noise spike during any time period equals the probability of a target spike belonging to the target spatio-temporal pattern.

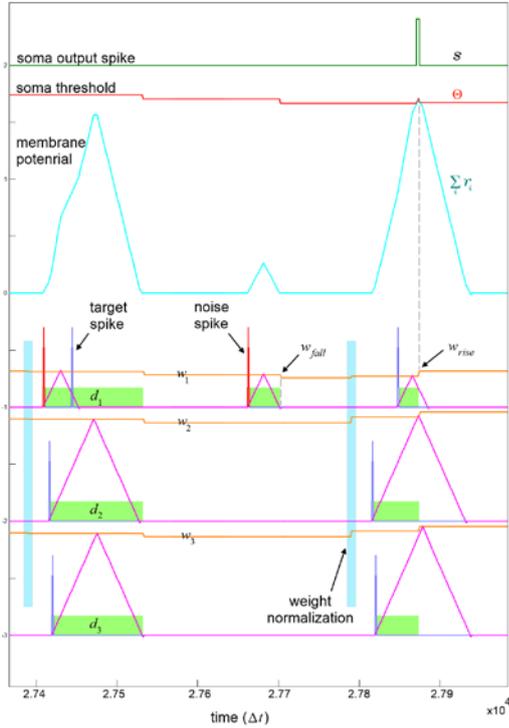

Figure 6. Synaptic weight adaptation in a neuron with three synaptic inputs. Input channel one is noisy (SNR=1:1) while the other two channels are noise free. Input spikes trigger both the kernel $r_i(t)$ and the weight update flag $d_i(t)$. The output spike $s(t)$ causes a rise in the synaptic weight of all synapses while the return of the membrane potential to zero induces a fall in the activated synapses. Also note the drop in threshold at t~2.77e4. Although noise spikes themselves typically do not cause output spikes, their presence at low levels can sometimes increase the likelihhood of a postsynaptic spike by reducing the neuron's threshold.

Noise spikes, being uncorrelated with the target pattern and with each other, typically arrive on their own or in such a way that their EPSPs are not enough to push the membrane potential $\Sigma r_i(t)$ past the threshold $\Theta(t)$ to cause a postsynaptic output spike. Such noise spikes do however reduce their respective synaptic weight by $w_{fall}$. This in turn causes an *increase* in the synaptic weight of less noisy channels when all channels are subsequently renormalized.

In the case where a clean target spatio-temporal pattern arrives without any neighboring noise, all the synaptic weights are increased equally by $w_{rise}$. However after the subsequent normalization the weights get recalibrated downward with a small bias in favor of the weakest synapses receiving more relative weight gain.

In the worst case scenario the noise spike(s) arrives just slightly ahead of a target spike. This disruption can be enough for a target pattern to be missed, causing every synaptic weight to decrease. Here again, the subsequent normalization means that the weakest synapses end up losing slightly more weight. This bias favoring greater change for the weaker synapses is in fact desirable since it produces inertia in the system by giving stronger synaptic weights with greater resistance to change.

The combined result of these changes is that synapses that only receive target input spikes accumulate higher and higher weights while synapses with greater noise have their weights pushed down. Over time the force pushing a synaptic weight up (postsynaptic spikes following presynaptic spikes), comes into balance with the forces pushing the weight down (presynaptic noise spikes and weight normalization). If homogenous Poisson processes generate the noise across the different channels, the synaptic weights converge to a steady state value which encodes the relative signal to noise ration of each afferent as shown in Figure 7.

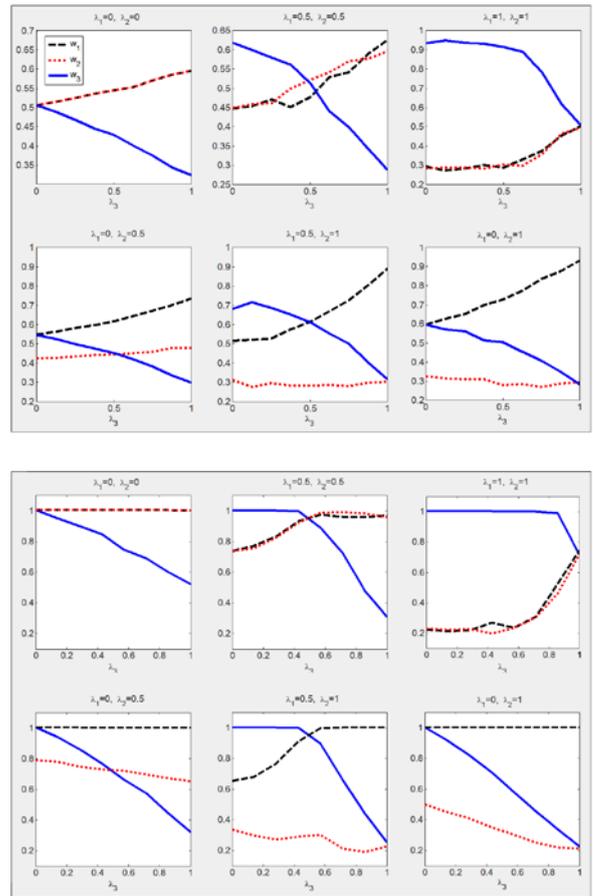

Figure 7. Synaptic weights as a function of noise. A sweep through the noise space was performed for the three synapse neuron where the neuron was receiving a random spatio-temporal target pattern corrupted with varying level of Poisson noise for each channel. In all panels the Poisson noise rate across the first two channels, $\lambda_1$ and $\lambda_2$, was kept constant while $\lambda_3$ was swept from SNR=0 to SNR=1 and the average steady state value of the synaptic weights, $w_1$, $w_2$ and $w_3$ are plotted. (top) Synaptic responses with $\Sigma w_i(t)$ used as normalization signal. (bottom) Synaptic responses with max($w_i(t)$) used as normalization signal.

To demonstrate how the synaptic weights of a neuron evolve to their steady state over time, various noise environments are shown in Figure 7 for the case of a sixteen input neuron. Here, the rest of the neuron's signals have been removed to clearly show how the synaptic weights encode the relative level of noise over time. The plots also show the greater stability of the higher weights.

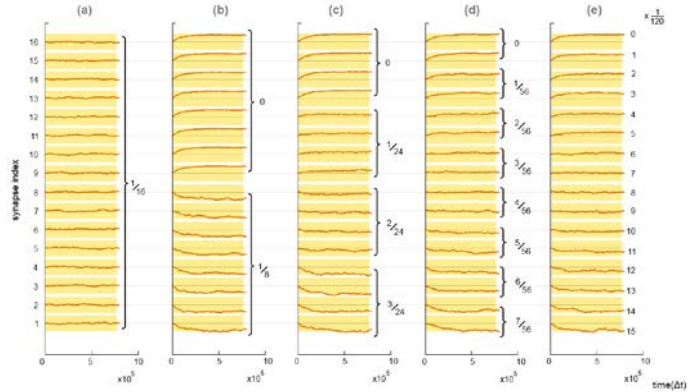

Figure 8. Time series plot showing the evolution of synaptic weights of a sixteen input neuron in a range of noise environments. The orange plots indicate the weight of each synapse while the yellow rectangles indicate the dynamic range of the synapse. As the figure demonstrates, the synaptic weights shift to compensate for the varying level of noise. (a) All the synapses having equal noise spike probability of 1/16 per target spike. (b) Half the channels are noise free while the other half have a 1/8 chance of receiving a noise spike for every target spike. (c) four groups of four synapses with varying noise levels (d) eight groups of two synapses with varying noise levels (e) Increasing noise levels with synapse number.

### B. Recognition performance of the Synaptic SNR encoding neuron on noise corrupted data

To quantify the recognition performance of the neuron under various noise regimes, the recognition error of the synaptic weight adapting neuron was measured against a neuron without synaptic weigh adaptation. Both neurons were presented with a random sequence one thousand patterns long populated by two random spatio-temporal patterns. For each test some of the input channels were noise corrupted at varying levels as indicated in Figure 8. The neurons were given one hundred presentations of this noisy randomized data stream within which to perform unsupervised learning of one of the two random patterns after which the pattern for which the neuron spiked most was designated as its target pattern. In the following nine hundred presentations the error in recognition was measured, defined as the number of missed target patterns plus false positive output spikes divided by the total number of target presentations. Figure 9 demonstrates the power of the SNR encoding synapse, where over a wide range of noise environments the neuron effectively removes all corrupting input noise and delivers near perfect unsupervised learning and recognition performance. An interesting feature of the neuron is the initial increase in error at the low noise level for the weight adapting neuron, where a minimum noise threshold must be reached to trigger the weight adaptation system. For these tests the $w_{rise}/w_{fall}$ ratio was deliberately chosen to clearly illustrate this behavior. This initial rise can be brought down by choosing a larger $w_{fall}$ term making the neuron more aggressive in terms of shutting down noisy afferents.

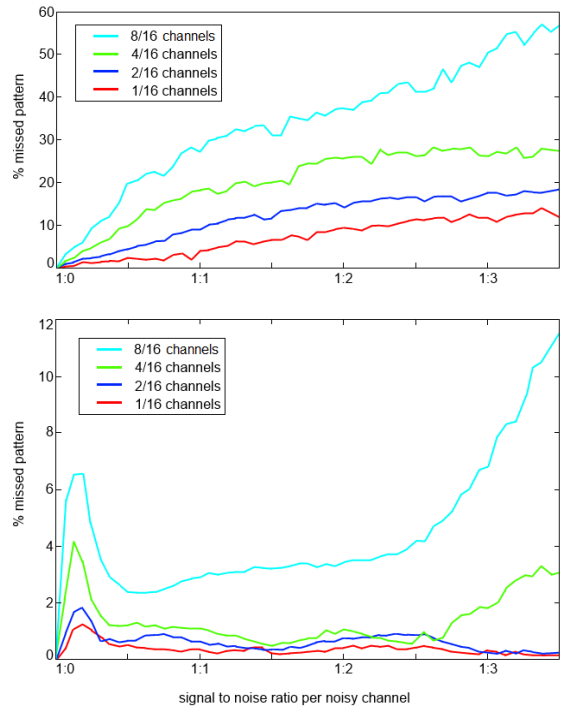

Figure 9. Enhanced recognition performance via synaptic signal to noise ratio encoding. (top) Recognition error as a function of increasing noise in a kernel adapting neuron with static synaptic weights. The four plots demonstrate decreasing performance both as the number of noisy channels are multiplied and as the SNR per noisy channel deteriorates. Note that as long as the noise corrupted channels are few in number the static SKAN can provide a moderate level of unsupervised recognition performance. (bottom) The same noise regime being applied to the same neuron this time with the dynamically adaptive synaptic weights (note the change in scale for the vertical axis). After perfect performance in the noiseless environments the error rates raise rapidly (SNR = 1:0 – 1:0.25). The reason for this initial rise in error is that the relative level of noise is simply too low to trigger the neuron's weight adaptation system such that the recognition profile is almost the same as for the static weighted neuron. As the noise level increases the neuron's SNR encoding system switches off the noisy channels and the recognition performance returns to near perfect. At very high noise levels (SNR < 1:3), the error rate begins to rise again, this time because the neuron's learning of its "target pattern" during the unsupervised learning period begins to deteriorate.

### C. Example Application: Unsupervised Feature Learning Using a Camera with Noise Corrupted Pixels

Cameras can often suffer from noisy pixels and experimental or neuromorphic cameras are specially prone to this problem. Cameras such as the event based DVS camera [21] can suffer from faulty pixels which generate noisy streams of pulses where there should be no activation and this can have a detrimental effect on an upstream recognition system. Such faults can require on-going examination of the camera by an expert user in order to detect and remove such noise corrupted pixels. Here a synaptic SNR encoding neuron is particularly useful in being able to simultaneously perform both the noise removal and the online unsupervised learning task without any expert intervention.

To demonstrate this capability the neuron was presented with the subset of all the zeros in the MNIST dataset [22]. The pixels in the dataset images were directly mapped to the synaptic input of the neuron. However since SKAN receives spike input the pixel values need to be converted to spike. The most straight forward approach, which was used here, involved simply mapping intensity to spike latency, with the brightest pixels arriving first and the darkest arriving last. This transformation was performed by a simple one to one mapping, however, neuromorphic approaches such as use of distributed integrate and fire neurons can also be used to convert real value signals to spike times [23].

To simulate the faulty camera some of the image pixels were corrupted with random levels of noise (1:1 < SNR < 1:3) as shown in Figure 10 (a). Here the design choice regarding zero weights referred to in section III B was implemented, where once a synaptic weight reached zero value, the synapse, and therefore the pixel, was disabled. This resulted in a system where the noise corrupted pixels were all disabled after at most 189 training images (see Figure 10 (d), leaving only the noise free channels for the kernels to train on and generating the receptive field of the neuron shown in Figure 10 (b). Note that the aggressiveness of pixel removal system can be controlled via the $w_{rise}/w_{fall}$ ratio.

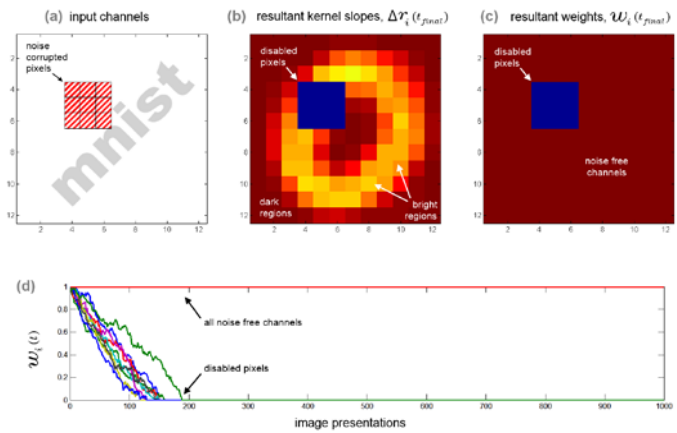

Figure 10 Developing the receptive field and SNR encoding weights for the noise corrupted handwritten zero digits of the MNIST dataset using a single neuron. (a) The input space with corrupted pixels highlighted. (b) The final receptive field of the neuron after exposure to the MNIST zeros. Pixels with higher probability of being dark are more likely to generate late spikes which a correctly trained neuron should encode in the form of narrow kernels or high kernel slopes $\Delta r_i(t)$. Conversely, pixels more likely to be bright should be encoded by lower kernel slopes, as is seen. As a result, the further an input image is from this 'model' of a zero, the weaker the response of the neuron to the image. (c) The final synaptic weights of the neuron showing the disabled pixels. (d) The evolution of the synaptic weight over time. The neuron correctly weighted all the equally noiseless pixels equally high while weakening the weight of the noisy channels until they reached their minimum value at which point they are disabled.

These disabled synapses shown in Figure 10 can potentially be reused, making SNR encoding synapses not only useful in terms of enhancing the performance of downstream signal processing systems as demonstrated in Figure 9, but in reconfigurable systems could also enable the reallocation of the valuable synaptic resources to other tasks. This would allow optimal hardware use in the context of the noise present in the sensors or even in the environment. Finally while

## V. CONCLUSION

A synapto-dendritic kernel adapting neuron, together with a simplified STDP update rule and a novel hardware efficient synaptic weight normalization system were combined produce neurons capable of synaptic encoding of afferent signal to noise ratio while performing unsupervised learning and recognition. The enhanced recognition performance and potential application of the neuron was shown in a visual feature learning task.